\documentclass{article}

    \usepackage[preprint]{neurips_2024}

\usepackage{caption}
\usepackage{tabularray}
\usepackage[utf8]{inputenc} 
\usepackage[T1]{fontenc}    
\usepackage{hyperref}       
\usepackage{url}            
\usepackage{booktabs}       
\usepackage{amsfonts}       
\usepackage{nicefrac}       
\usepackage{microtype}      
\usepackage{xcolor}         
\usepackage{graphicx}

\usepackage{natbib}
\usepackage{verbatim}
\usepackage{inconsolata}
\usepackage{tabularx}
\usepackage{float}
\usepackage{algorithm}
\usepackage{algorithmic}
\usepackage{multirow}
\usepackage{listings}
\usepackage{xcolor}
\usepackage{subcaption}
\usepackage{amsmath}
\usepackage{amssymb}
\definecolor{lightgray}{rgb}{0.95, 0.95, 0.95}
\definecolor{darkgray}{rgb}{0.4, 0.4, 0.4}
\definecolor{backcolour}{rgb}{0.95,0.95,0.92}
\definecolor{myblue}{rgb}{0.2, 0.4, 0.8} 
\definecolor{mygreen}{rgb}{0.2, 0.6, 0.2} 

\usepackage{amsmath,amssymb,amsfonts}

\title{Large Language Models as Code Executors: An Exploratory Study }

\author{
  Chenyang Lyu$^1$\thanks{Work done while at MBZUAI} \quad 
  Lecheng Yan$^2$ \quad 
  Rui Xing$^{1,3}$ \quad
  Wenxi Li$^{4,5}$ \\ 
  \textbf{Younes Samih}$^6$ \quad
  \textbf{Tianbo Ji}$^7$ \quad
  \textbf{Longyue Wang}$^8$ \\
  $^1$Mohamed bin Zayed University of Artificial Intelligence \\
  $^2$Xinjiang University
  $^3$University of Melbourne 
  $^4$Tsinghua University \\
  $^5$Shanghai Jiao Tong University
  $^6$IBM Research
  $^7$Nantong University
  $^8$Alibaba \\
  \texttt{lyuchenyang.dcu@gmail.com}
}

\begin{document}

\maketitle

\begin{abstract}

The capabilities of Large Language Models~(LLMs) have significantly evolved, extending from natural language processing to complex tasks like code understanding and generation. We expand the scope of LLMs' capabilities to a broader context, using LLMs to \textit{execute} code snippets to obtain the output. This paper pioneers the exploration of LLMs as code executors, where code snippets are directly fed to the models for execution, and outputs are returned. We are the first to comprehensively examine this feasibility across various LLMs, including OpenAI's o1, GPT-4o, GPT-3.5, DeepSeek, and Qwen-Coder. Notably, the o1 model achieved over 90\% accuracy in code execution, while others demonstrated lower accuracy levels. Furthermore, we introduce an Iterative Instruction Prompting~(IIP) technique that processes code snippets line by line, enhancing the accuracy of weaker models by an average of 7.22\% (with the highest improvement of 18.96\%) and an absolute average improvement of 3.86\% against CoT prompting (with the highest improvement of 19.46\%). Our study not only highlights the transformative potential of LLMs in coding but also lays the groundwork for future advancements in automated programming and the completion of complex tasks.
\end{abstract}

\section{Introduction}

\begin{figure}[htbp]
    \centering
    \includegraphics[width=0.9\linewidth]{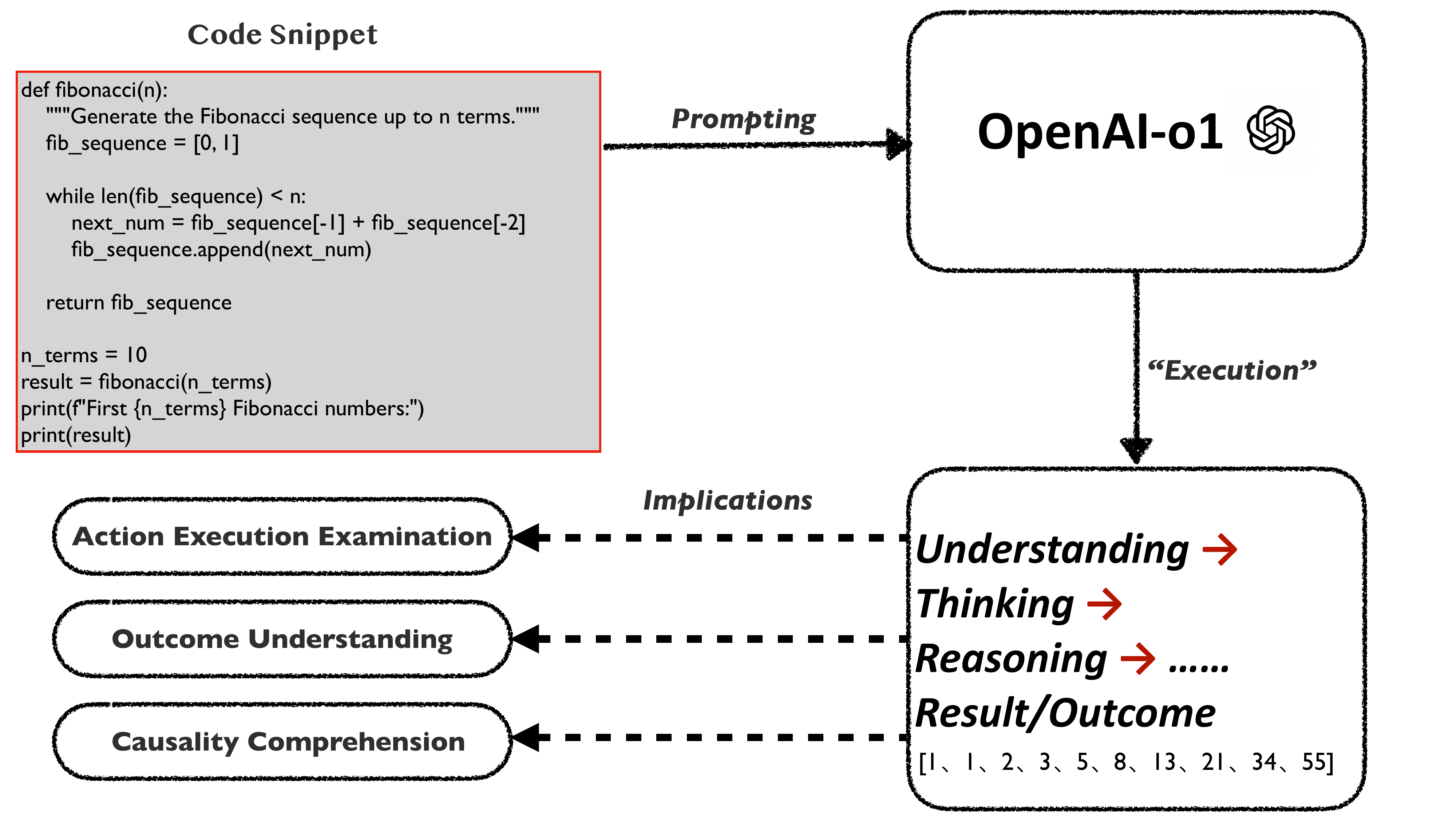}
    \caption{An illustration to the use of LLMs for code execution as a proxy for evaluating execution competence as well as the implications for understanding outcomes, and progressing towards real-world causality comprehension.}
    \label{fig:intro}
\end{figure}
The rapid advancement of Large Language Models~(LLMs)~\citep{brown2020language_gpt3,openai2023gpt4} has made a transformation of capabilities across diverse domains, ranging from language translation to creative writing~\citep{wang2023document,bang2023multitask_chatgpt_eval,bai2023qwen}. These models, with their remarkable ability to understand and generate human-like text, have found applications that extend well beyond traditional natural language processing tasks such as code understanding and generation~\citep{chen2021evaluating-codex-gpt3,gao2023makes,zhuo2024bigcodebench}. In the area of programming, LLMs have been predominantly utilized for code generation, aiding developers by suggesting code snippets or completing partially written scripts~\citep{wang2024systematicevaluationlargecode}. This utility has significantly enhanced productivity and coding efficiency by providing real-time assistance and reducing the cognitive load on developers~\citep{jiang2024surveylargelanguagemodels}.

Despite these advancements, the exploration of LLMs as code executors remains a less explored area. The ability to not only generate but also execute code opens up a plethora of possibilities, including automated debugging, real-time code validation, and the development of intelligent programming assistants. More importantly, this links to a broader context and higher-level gold of using LLMs to execute and complete complex actions and plans and even for causality understanding in real world~\citep{wang2023codet5opencodelarge,kambhampati2024position}. This paper is pioneering in its approach, as it is the first to systematically examine the feasibility of employing LLMs to execute code directly, providing immediate feedback and results—an evolutionary leap from mere code suggestion to active code execution. An straightforward illustration of this idea is shown in Figure~\ref{fig:intro}, which describes the central concept of using LLMs like OpenAI's o1 to execute code, emphasizing their role in evaluating execution competence, understanding outcomes, and progressing towards real-world causality comprehension. By processing code snippets, LLMs can serve as a benchmark or proxy for assessing their ability to perform tasks and demonstrate understanding of the results and implications of actions. This approach aims to enhance problem-solving capabilities and extend the application scope of LLMs beyond algorithmic tasks, ultimately paving the way for models to grasp the causality of real-world actions.

In our study, we evaluated various LLMs, such as OpenAI's o1~\footnote{\url{https://openai.com/index/openai-o1-system-card/}}, GPT-4o~\footnote{\url{https://openai.com/index/gpt-4o-system-card/}}, GPT-3.5~\citep{ouyangtraining-instructGPT}, DeepSeek~\citep{guo2024deepseekcoderlargelanguagemodel}, and Qwen-Coder~\citep{hui2024qwen25codertechnicalreport}, to assess their performance as code executors. Our experiments reveal that the latest OpenAI o1 model achieves a remarkable execution accuracy of over 90\%, setting a new benchmark in this field. In contrast, other models demonstrate significantly lower accuracy, often falling below the 50\% threshold. This disparity highlights the need for innovative techniques to boost the performance of less accurate models.

To address this challenge for most of the LLMs, we propose an Iterative Instruction Pprompting technique~(IIP) inspired by Chain-of-Thoughts~\citep{wei2022chain_cot} and Tree-of-Thoughts~\citep{yao2024tree_tot} prompting. This method involves feeding code snippets into LLMs line by line, allowing the models to process and execute each segment individually before generating the final output. This approach not only enhances the comprehension and execution accuracy of the models but also results in an average 7.22\% improvement for those with lower baseline performance and an improvement of 3.86\% against Chain-of-Thoughts prompting~\citep{wei2022chain_cot,kojima2022large_cot}. We also analyse the effect of various factors such as coding type, lines of code snippets and the computational complexity to the performance of LLMs. This exploration and analysis could potentially transform current coding practices by enabling more robust and reliable automated code execution.

In this work, we aim to further expand the utility and functionality of LLMs. By demonstrating the potential of LLMs as code executors, we lay the groundwork for future exploration into automated software development, where intelligent programming assistants could revolutionize how code is written, executed, tested, and deployed. This paper seeks to provide fresh insights and ispire further research into leveraging LLMs for more sophisticated programming tasks and complex tasks for action execution and causality comprehension.

\section{LLMs as Code Executors}

In this section, we outline the methodology employed to investigate the capabilities of LLMs as code executors. Our approach is designed to systematically evaluate how effectively LLMs can execute code snippets and return accurate outputs, an interesting application that extends their use beyond traditional code generation tasks. This involves the collection of diverse code snippets and the careful design of prompts that guide the models in executing code. By doing so, we aim to uncover insights into the operational dynamics of LLMs when tasked with direct code execution and to identify strategies that enhance their performance. The following subsections detail the processes of code snippet collection and prompt design, which form the foundation of our experimental framework.

\begin{figure}
    \centering
    \includegraphics[width=0.9\linewidth]{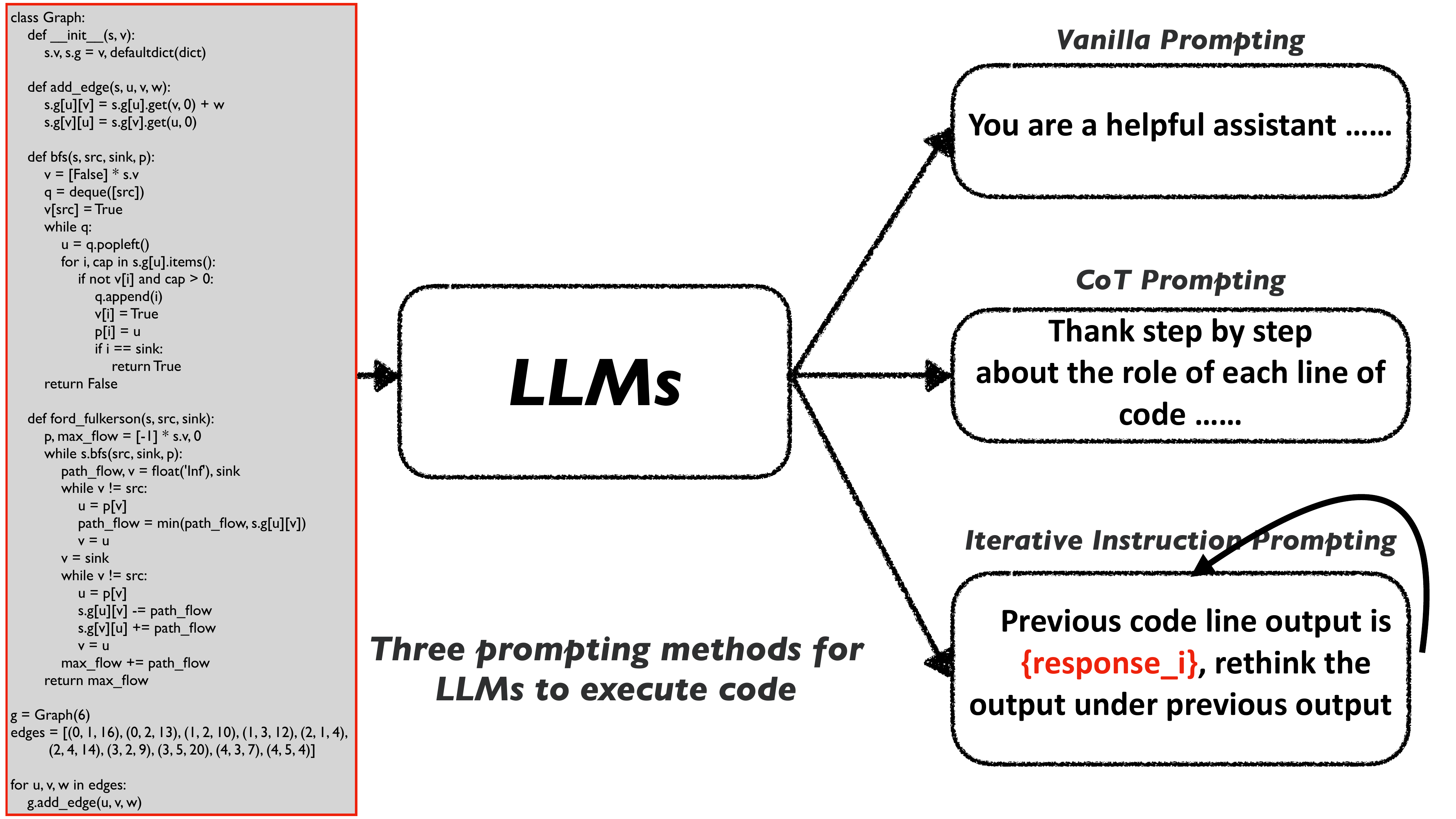}
    \caption{Comparison of prompting methods for LLM code execution: vanilla prompting for general guidance, Chain-of-Thought~(CoT) prompting for step-by-step analysis, and iterative instruction prompting for refining outputs based on prior responses.}
    \label{fig:method}
\end{figure}

\begin{table}[H]
\centering
\caption{The amount of the main different question types in the dataset, where DP is \textit{Dynamic Programming}.}
\resizebox{\linewidth}{!}{%
\begin{tabular}{lcccccccccc}
    \toprule
    Source & Array & Greedy & DP & String & Math & Binary Search & Stack & Heap & Recursion & Sorting \\
    \midrule
    English & 61 & 45 & 25 & 25 & 24 & 22 & 17 & 15 & 15 & 14 \\
    Chinese & 60 & 32 & 27 & 19 & 18 & 16 & 15 & 13 & 11 & 10 \\
    \bottomrule
\end{tabular}%
}
\end{table}
\subsection{Code Snippets Collection}
We collected code snippets from Leetcode~\footnote{\url{https://leetcode.com}}, including 100 examples in both Chinese and English respectively. The platform provides data such as problem description, test cases, standard solutions and problem types corresponding to each problem, and we collect the matching data manually and then analyze it, and finally, the format of each of our metadata is as follows:
\begin{enumerate}
    \item Problem Descriptions. For each code snippet, we provide a detailed description of the problem it aims to solve.
    \item Input-Output Examples. Each question is accompanied by the corresponding input data and the corresponding expected output, which is used as the LLMs evaluation data
    \item Standard solution. We include a standard solution for each problem, describing in detail the idea of solving each problem in order to provide a cross-reference to the collected code snippets.
    \item Solution Code. For the standard solution, we have collected the corresponding Python, Java, C, C\texttt{++} four different problem solving code, so as to provide a perfect solution for each code fragment.
    \item Problem Type. We have collected the corresponding types of each problem, such as strings, arrays, sorting, math, etc., for targeted analysis and application to understand how LLMs perform in different types of
    \item Problem Difficulty and Human Pass Rate. As an objective response to the difficulty of the problem versus the reality of the situation, indicating the percentage of humans who successfully solved each problem on the LeetCode platform.
    \item Other. By re-analyzing the existing data, we record the number of lines of code and time complexity of each solution.
\end{enumerate}

In the end, we collected 100 examples in English and 100 examples in Chinese, in total 200 code snippets.

\subsection{Prompting Designation}
As illustrated in Figure \ref{fig:method}, three approaches are compared: Vanilla Prompting~(VP), Chain-of-Thought~(CoT) prompting, and our proposed Iterative Instruction Prompting~(IIP). VP serves as a basic interaction model, providing general assistance without specific guidance. In contrast, CoT prompting facilitates a more detailed analysis by encouraging the model to consider the role of each line of code. Finally, IIP makes LLMs to receive code snippets line by line and builds upon previous outputs, allowing the model to refine its predictions based on earlier results. This comparison highlights the progressive enhancement of LLM capabilities in understanding and executing complex code tasks. Below is the vanilla prompting we employed without sophisticated design:

\begin{lstlisting}
~System~:
    You are a helpful assistant
*User*:
    This is our python code:
    @{python_code}@
    what is the result/output of this code if the input is @{input_data}@?
\end{lstlisting}

Moreover, in order to clarify the task of LLMs as code executors, while considering improving their performance in this task, we emphasize their role in considering each line of code, ultimately using the following CoT prompt:

\begin{lstlisting}
~System~:
    You are a programming expert, good at python code, especially algorithmic
    problems, please think step by step about the execution of the code steps, 
    think about the role of each line of code, get the result.
*User*:
    This is our python code:
    @{python_code}@
    what is the result/output of this code if the input is @{input_data}@?
\end{lstlisting}

Furthermore, we propose an approach that iteratively prompts LLMs with the code snippets line by line, which allows LLMs to think about the output of the previous code line as the input to the next code line, so we use the following prompt:

\begin{lstlisting}
~System~:
    You are a programming expert, good at python code, especially algorithmic
    problems, please think step by step about the execution of the code steps, 
    think about the role of each line of code, get the result.
*User*:
    begin:
    This is our python code:
    @{python_code}@
    what is the result/output of this code if the input is @{input_data}@?
    Now please output the execution process of each line of code in the code,
    and output the variable results after each step of the code while executing
    each line of code.

    process:
    My previous analysis process is as follows:
    @{response_i}@,
    As for the previous analysis process, please help me rethink the output under
    this input, and also output the execution process of each line of code in the
    code, output the variable results after each step of code while executing each
    line of code.

    end:
    My previous analysis process is as follows:
    @{response_n}@,
    As for the previous analysis process, please help me rethink the output under
    this input, and also output the execution process of each line of code in the
    code, output the variable results after each step of code while executing each
    line of code, and finally output the most possible result.
\end{lstlisting}

The iterative code execution process involves several steps to compute outputs from a set of code lines and refine these outputs iteratively. Below is a detailed description of this approach:

Let \( C = \{c_1, c_2, \ldots, c_n\} \) be the set of code lines, and let \( I_0 \) represent the initial input data. We aim to compute the output for each line of code and update our inputs iteratively.

The process begins with the generation of the output \( O_i \) for each line \( c_i \) in \( C \). This is achieved by applying a function \( f \), which takes the current line of code and the previous input as arguments:

\[
O_i = f(c_i, I_{i-1})
\]

After the computation, the input for the subsequent line is updated with the output of the current line. Thus, the input update rule is given by:

\[
I_i = O_i
\]

The iterative refinement process involves re-evaluating the outputs. For each iteration \( k \), the outputs are recalculated to enhance accuracy:

\[
O_i^{(k+1)} = f(c_i, I_{i-1}^{(k)})
\]

Finally, after \( n \) iterations, the process converges to a final output \( O_{\text{final}} \), which is represented as:

\[
O_{\text{final}} = O_n^{(k)}
\]

This method ensures that each line's execution is informed by the previous computations, providing a robust framework for code execution with iterative refinement.

\section{Evaluation Results}
In this section, we present the evaluation results of various LLMs on the code snippets we collected.

\subsection{Evaluated LLMs}

\begin{itemize}
    \item \textbf{GPT-3.5:} GPT-3.5~\citep{ouyangtraining-instructGPT} improves on GPT-3~\citep{brown2020language_gpt3} with enhanced language capabilities, supporting zero-shot and few-shot learning. The GPT-3.5 turbo variant balances cost, latency, and quality. The model name used is \textit{gpt-3.5-turbo-0125}.

    \item \textbf{GPT-4:} GPT-4~\citep{openai2023gpt4} is known for its advanced language understanding and generation capabilities, which is a further improved version of GPT-3.5. In our experiments, we test \textit{gpt-4-turbo-2024-04-09}.

    \item \textbf{GPT-4o:} GPT-4o can work on text and image processing with excellent multimodal capability, supporting 50 languages with enhanced memory capabilities for context retention. The model we used is \textit{gpt-4o-2024-08-06} and \textit{gpt-4o-mini-2024-07-18}.

    \item \textbf{OpenAI's o1:} The OpenAI-o1 model has excellent complex reasoning capabilities, especially in complex tasks such as coding and scientific research. The model name used is \textit{o1-preview-2024-09-12}.

    \item \textbf{DeepSeek-Coder:} DeepSeek-Coder~\citep{guo2024deepseekcoderlargelanguagemodel}, an open-source Mixture-of-Experts model, matches GPT4-Turbo in code tasks. Pre-trained with an additional 6 trillion tokens, it supports 338 programming languages and extends context length to 128K.

    \item \textbf{Qwen-2.5-Coder:} Qwen-2.5-Coder~\citep{hui2024qwen25codertechnicalreport} based on Qwen LLMs~\citep{bai2023qwen} is optimized for coding, supporting 128K tokens and 92 programming languages. Trained on 5.5 trillion tokens, it excels in code generation and reasoning. We also use Qwen-2.5-72B in our experiments.
\end{itemize}

\subsection{Experimental Setup}
Based on the data we collected above, we extract the python code with test cases for each metadata, embed it in the set prompt, set each test case to be asked twice as LLMs input and record the corresponding output; for our IIP approach, we take the last LLMs replies and embed it in the next prompt.

\subsection{Results}

\paragraph{Main Results}
\begin{table}[h]
    \centering
    \caption{Experimental results for LLMs on CN and EN data, where the highest performance is marked as \textbf{bold} and the second best accuracy is marked with \underline{underscore}.}
    \resizebox{\linewidth}{!}{
    \begin{tabular}{lcccccccc}
        \toprule
        & \textbf{GPT-3.5} & \textbf{GPT-4} & \textbf{GPT-4o} & \textbf{GPT-4o-mini} & \textbf{o1-Preview} & \textbf{Qwen-Coder} & \textbf{Qwen-72B} & \textbf{DeepSeek-Coder} \\
        \midrule
        \textbf{CN} & 32.2 & 49.7 & \underline{66.7} & 52.3 & \textbf{93.5} & 20.4 & 58.8 & 60.0  \\
        \textbf{EN} & 40.0 & 61.3 & \underline{73.8} & 64.3 & \textbf{96.1} & 23.8 & 73.4 & 70.5  \\
        \bottomrule
    \end{tabular}
    }

    \label{tab:main_results}
\end{table}

The experimental evaluation was conducted to assess the performance of various LLMs in executing code snippets, sourced both from CN (Chinese) and EN (English) contexts. The LLMs tested include GPT-3.5, GPT-4, GPT-4o, GPT-4o-mini, o1-Preview, Qwen-Coder, Qwen-72B, and DeepSeek-Coder. Each model was tasked with executing code snippets embedded with comments in their respective languages, providing a comprehensive overview of their capabilities across different linguistic and syntactic environments.

The results, as presented in Table \ref{tab:main_results}, highlight several key insights. Notably, the latest OpenAI o1-Preview model consistently outperformed the others, achieving an accuracy of 93.5\% for CN and 96.1\% for EN, suggesting highly excellent ability for this code execution task and a robust capability to handle code execution across diverse linguistic inputs. This indicates the o1's superior ability to process and understand the nuances of code comments and structure, further confirming the capability of solving complex tasks of o1 model. In contrast, models such as GPt-3.5, Qwen-Coder and Qwen-72B demonstrated lower performance, with accuracy around 20\% to 60\%, which substantially lag behind o1's performance. This suggests that these models may lack sufficient training or optimization for code execution tasks, particularly in handling complex code structures or understanding context from comments.

Another interesting observation is the performance disparity between CN and EN across models. While most models showed better performance with English code snippets, the margin varied, indicating potential biases or limitations in handling code semantics when embedded in non-English contexts. This highlights the need for further refinement in multilingual code execution capabilities.

\paragraph{Effect of prompt type}

\begin{table}[h]
\centering
\caption{Comparison of average accuracy for different prompt types (EN and CN), the highest performance and improvements are in bold.}
\begin{tabular}{llccc}
\toprule
\textbf{Model} & \textbf{Source} & \textbf{Vanilla} & \textbf{CoT} & \textbf{IIP} \\
\midrule
\multirow{2}{*}{GPT-3.5} & CN & 27.05 & 32.18 (+5.13) & \textbf{33.92~(+6.87)} \\
 & EN & 31.92 & 40.00 (+8.08) & \textbf{40.46~(+8.54)} \\
\midrule
\multirow{2}{*}{Qwen-2.5-Coder} & CN & 18.63 & 20.42~(+1.79) & \textbf{33.70 (+15.07)} \\
 & EN & 24.33 & 23.83~(-0.50) & \textbf{43.29~(+18.96)} \\
\midrule
\multirow{2}{*}{Qwen-2.5-72B} & CN & 55.58 & 58.79~(+3.21) & \textbf{63.50~(+7.92)} \\
 & EN & 69.45 & 73.38~(+3.93) & \textbf{76.13~(+6.68)} \\
\midrule
\multirow{2}{*}{Deepseek-Coder} & CN & 55.53 & \textbf{60.04~(+4.51)} & 54.03~(-1.50) \\
 & EN & 69.75 & \textbf{70.50~(+0.75)} & {64.96~(-4.79)} \\
\bottomrule
\end{tabular}
\label{tab:results_prompt_type}
\end{table}

This experiment examines the impact of different prompting strategies on the performance of various LLMs in executing code snippets. The study evaluates three types of prompts: vanilla, CoT~\citep{wei2022chain_cot,kojima2022large_cot}, and Iterative Instruction Prompting~(IIP), across both EN and CN datasets. The LLMs we used include GPT-3.5, Qwen-2.5-Coder, Qwen-2.5-72B, and Deepseek-Coder.

The Vanilla Prompt serves as the baseline, representing the simplest form of instruction. The CoT prompt involves role-playing as an expert, providing the model with additional context and guidance as well as step-by-step thinking and reasoning. IIP, or iterative prompting, involves breaking down the code snippets line by line to feed into LLMs to improve comprehension and execution. This experimental setup aims to discern how each prompting technique influences model accuracy in code execution tasks.

The results shown in Table~\ref{tab:results_prompt_type} demonstrate varying degrees of effect across models and languages when transitioning from vanilla to more sophisticated prompt types. Notably, the IIP consistently yields the highest accuracy improvements, particularly in the CN dataset except for Deepseek-Coder model (IIP leads to slightly lower performance). For instance, Qwen-2.5-Coder shows a significant accuracy increase from 18.63\% with vanilla prompts to 33.70\% with IIP, highlighting the effectiveness of iterative prompting in enhancing model performance through step-by-step guidance. In the case of GPT-3.5, both CoT and IIP prompts lead to improvements in the CN dataset, with IIP offering a slightly higher boost. However, its performance remains relatively stable in the EN dataset, suggesting that the model may already be optimized for English code snippets, and further prompting variations have limited impact. Interestingly, Deepseek-Coder shows a decrease in accuracy with IIP prompting in both CN and EN datasets. This suggests that while iterative prompting benefits some models, its efficacy may depend on specific LLMs.

\paragraph{Performance across different question types}

\begin{figure}
    \centering
    \includegraphics[width=0.9\linewidth]{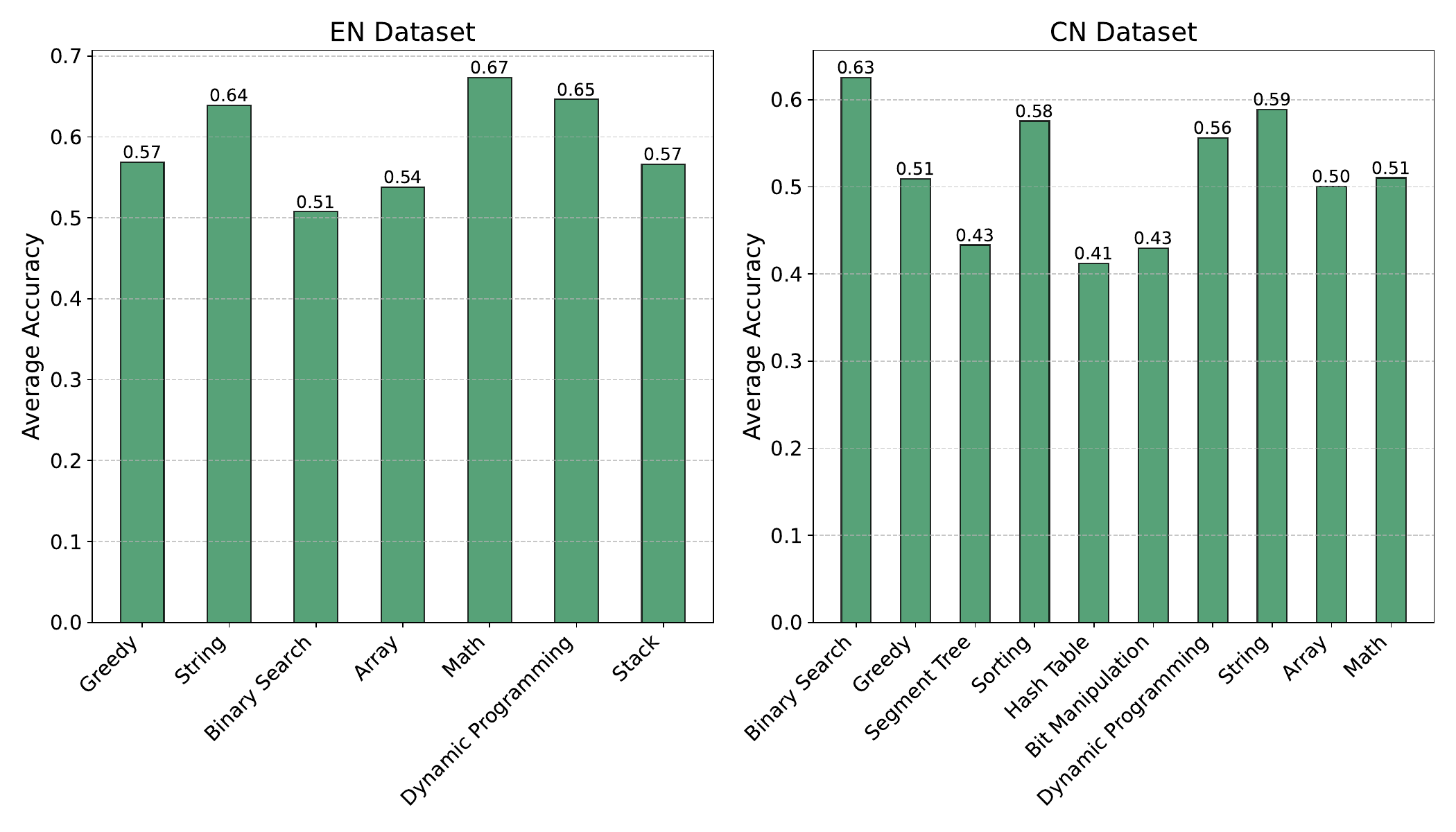}
    \caption{Split of average accuracy of all LLMs across different categories for both EN and CN code.}
    \label{fig:question_type_results}
\end{figure}

We further analyze the average performance of LLMs across various selected problem categories~(categories are not disjoint sets) that appeared more than ten times in our dataset of 100 questions. The categories include dynamic programming, array, segment tree, sorting, bit operation, binary search, greedy algorithms, hash table, mathematics, and string manipulation. As shown in Figure~\ref{fig:question_type_results}, the models achieved the highest average accuracy of 0.63 in binary search questions, indicating a strong proficiency in handling structured search algorithms likely due to their deterministic nature. In contrast, bit manipulation and dynamic programming showed lower accuracies of 0.41 and 0.43, respectively, suggesting these areas challenge LLMs possibly due to the complex logic and recursive reasoning required.

Performance in array and string manipulation was moderate, with accuracies of 0.50 and 0.51, indicating that while LLMs handle basic operations, they struggle with more advanced cases. Interestingly, models performed well in mathematics and sorting problems, achieving accuracies of 0.59 and 0.58, which reflects their ability to leverage algorithmic thinking and numerical computation.
 
\paragraph{Relationship between model accuracy and human pass rate}

\begin{figure}[htbp]
    \centering
    \includegraphics[width=1\linewidth]{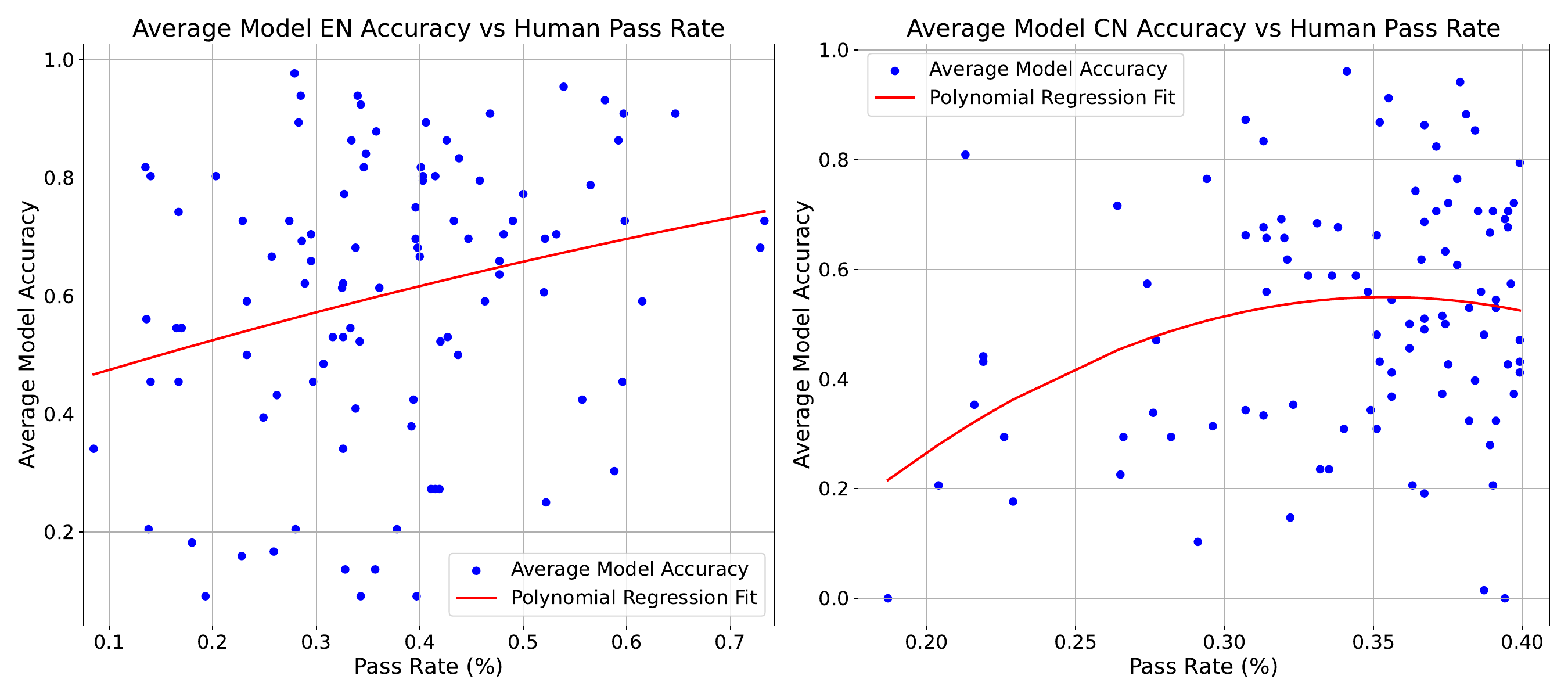}
    \caption{Relationship between human pass rates and model accuracy for EN~(left) and CN~(right) code snippets, the Spearman’s Correlation for EN and CN data are 0.25 and 0.17 respectively.}
    \label{fig:human_pass_rate}
\end{figure}

This section analyses the relationship between the average accuracy of all evaluated LLMs in Table~\ref{tab:main_results} and corresponding human pass rates of each coding question. The fit results are shown in Figure~\ref{fig:human_pass_rate}. Both EN and CN datasets show a positive correlation, indicating that tasks easier for humans generally yield higher model accuracy. The fit lines in the plots suggest that LLMs are more adept at solving problems with higher human pass rates, likely due to shared cognitive processes or data patterns. Notably, the EN dataset displays a steeper correlation, possibly due to more extensive training data or linguistic characteristics favoring English comprehension for the comments in code snippets.

\paragraph{Effect of Computational Complexity}

\begin{figure}[htbp]
    \centering
    \includegraphics[width=1.0\linewidth]{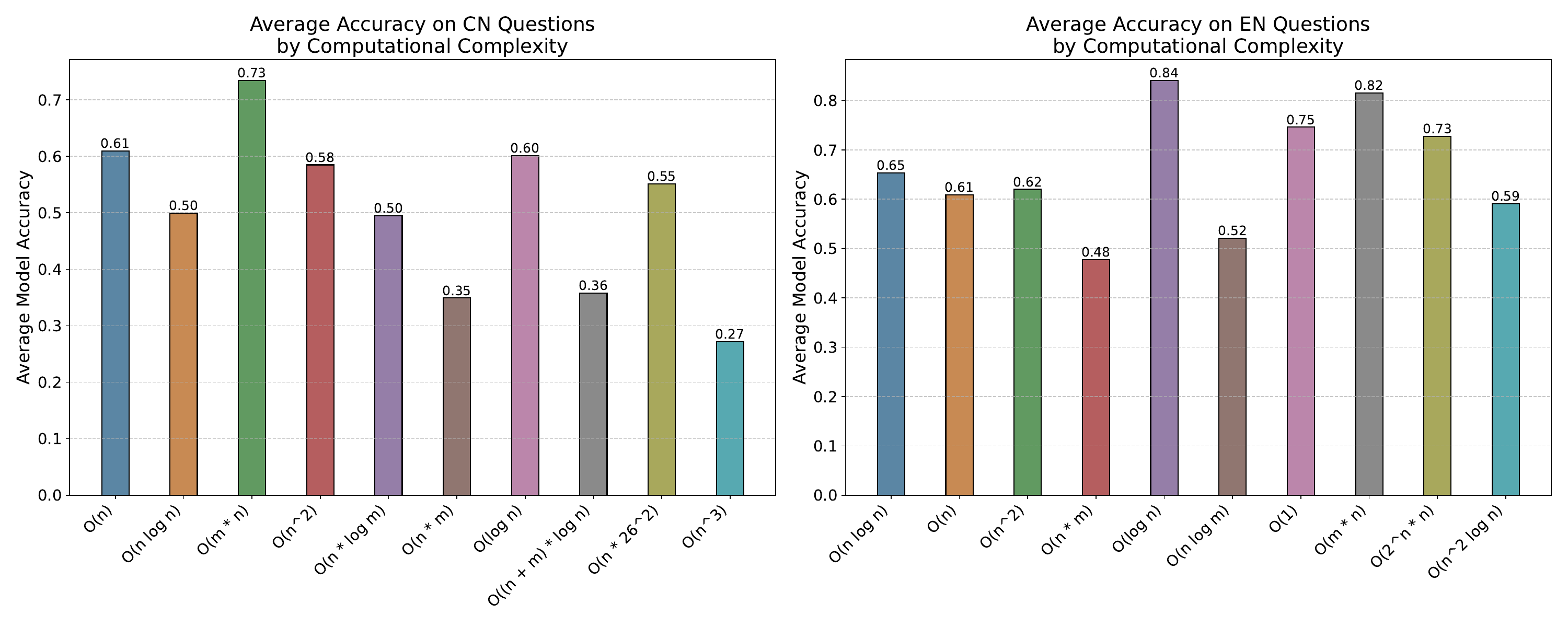}
    \caption{Average accuracy on CN and EN code snippets divided by corresponding computational complexity.}
    \label{fig:results_complexity}
\end{figure}

This section evaluates LLM performance on code snippets by analyzing average accuracy relative to computational complexity across CN and EN datasets. The complexities considered include \(O(n)\), \(O(n \log n)\), \(O(n^2)\), and others as shown in Figure~\ref{fig:results_complexity}, providing insights into model capabilities across varying algorithmic difficulties.

The resutls are shown in Figure~\ref{fig:results_complexity}. For CN questions, LLMs show strong performance in \(O(n \log n)\) tasks with an accuracy of 0.73, indicating proficiency in moderately complex problems like sorting. However, accuracy drops to 0.27 for \(O(n^3)\) tasks, highlighting challenges with highly complex operations. In EN questions, models achieve the highest accuracy of 0.84 for \(O(2^n \cdot n)\) tasks, suggesting strong capabilities in handling exponential growth problems Simpler complexities like \(O(n)\) yield moderate accuracies in both datasets, reflecting efficiency in straightforward tasks but indicating room for improvement in handling nested operations.

\paragraph{Effect of lines of code snippets}

\begin{figure}
    \centering
    \includegraphics[width=1.0\linewidth]{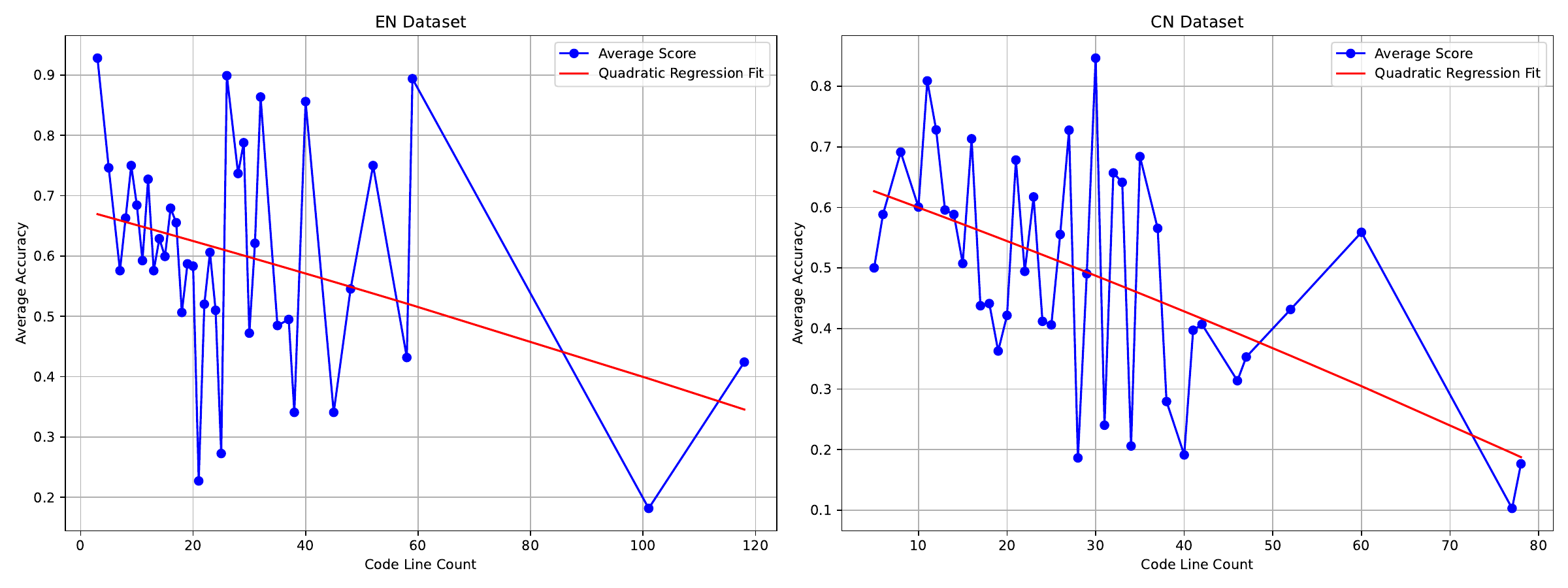}
    \caption{Overall regression analysis illustrating the impact of code snippet length on model accuracy for CN and EN datasets, the Spearman’s Correlation for EN and CN are -0.32 and -0.53 respectively.}
    \label{fig:code_line}
\end{figure}

We also the relationship between LLM accuracy and code snippet length in CN and EN datasets. As shown in Figure~\ref{fig:code_line}, quadratic regression analysis reveals a negative correlation between line count and accuracy for both datasets, indicating that longer code snippets generally reduce model performance. The CN dataset shows a steeper decline, suggesting greater challenges in handling complexity compared to the EN dataset, where the impact is less pronounced. These findings underscore the need to enhance LLMs' ability to manage code complexity, particularly in multilingual contexts, by developing more sophisticated code understanding mechanisms to improve robustness against variability in code length.

\subsection{Case Analysis}

We further analyze the performance of each model in our dataset. For example, question 91 of the English dataset: \textit{Booking Concert Tickets in Groups}. As a complex search problem, its standard code answer is more than one hundred lines. In the experiment, all LLM except OpenAI's o1 could not correctly handle the problem conditions and search and assign them, which proved o1's superior ability in dealing with complex problems.

\begin{lstlisting}[language=Python]
class Node:
    def __init__(self, start, end):
        self.s = start
        self.e = end
        self.left = None
        self.right = None
        self.total = 0 # for range sum query
        self.mx = 0 # for range max query
        
class SegTree:
    def __init__(self, start, end, val):
        
        def build(l, r):
            if l > r:
                return None
            if l == r:
                node = Node(l, r)
                node.total = val
                node.mx = val
                return node
            node = Node(l, r)
            m = (l + r) // 2
            node.left = build(l, m)
            node.right = build(m+1, r)
            node.mx = max(node.left.mx, node.right.mx)
            node.total = node.left.total + node.right.total
            return node
        
        self.root = build(start, end)
    
    # update the total remain seats and the max remain seats for each node (range) in the segment tree
    def update(self, index, val):
        
        def updateHelper(node):
            if node.s == node.e == index:
                node.total -= val
                node.mx -= val
                return
            m = (node.s + node.e) // 2
            if index <= m:
                updateHelper(node.left)
            elif index > m:
                updateHelper(node.right)
            node.mx = max(node.left.mx, node.right.mx)
            node.total = node.left.total + node.right.total
            return
            
        updateHelper(self.root)
        
    def maxQuery(self, k, maxRow, seats):
        
        def queryHelper(node):
            if node.s == node.e:
                # check if the row number is less than maxRow and the number of remains seats is greater or equal than k
                if node.e > maxRow or node.total < k:
                    return []
                if node.e <= maxRow and node.total >= k:
                    return [node.e, seats - node.total]
                # we want to greedily search the left subtree
    ........................... more  lines of code
\end{lstlisting}

Meanwhile, for question 85 in the Chinese dataset: \textit{Subarrays Distinct Element Sum of Squares II}. It is a combination of Segment Tree, Array, and Math, which contains numerical computation and remainder operation for large integers. Most LLMs fail in the remainder operation, while o1 completes all the steps perfectly, proving that o1 is better than other models in numerical processing and operations.

\begin{lstlisting}[language=Python]
class Solution:
    def sumCounts(self, nums: List[int]) -> int:
        n = len(nums)
        sum = [0] * (n * 4)
        todo = [0] * (n * 4)

        def do(o: int, l: int, r: int, add: int) -> None:
            sum[o] += add * (r - l + 1)
            todo[o] += add

        # o=1  [l,r] 1<=l<=r<=n
        def query_and_add1(o: int, l: int, r: int, L: int, R: int) -> int:
            if L <= l and r <= R:
                res = sum[o]
                do(o, l, r, 1)
                return res

            m = (l + r) // 2
            add = todo[o]
            if add:
                do(o * 2, l, m, add)
                do(o * 2 + 1, m + 1, r, add)
                todo[o] = 0

            res = 0
            if L <= m: res += query_and_add1(o * 2, l, m, L, R)
            if m < R:  res += query_and_add1(o * 2 + 1, m + 1, r, L, R)
            sum[o] = sum[o * 2] + sum[o * 2 + 1]
            return res

        ans = s = 0
        last = {}
        for i, x in enumerate(nums, 1):
            j = last.get(x, 0)
            s += query_and_add1(1, 1, n, j + 1, i) * 2 + i - j
            ans += s
            last[x] = i
        return ans % 1_000_000_007
    ........................... more  lines of code
\end{lstlisting}

In addition, for question 7 of the Chinese dataset: \textit{Alternating Groups III}.
The standard answer to this question lacks the definition of the tree-like structure given in the question. While the rest of the LLMs refused to answer or answered incorrectly due to the lack of conditions, o1 successfully deduced the conditions and answered correctly, which further proved its strong generalization ability.

\begin{lstlisting}[language=Python]
from sortedcontainers import SortedList

class FenwickTree:
    def __init__(self, n: int):
        self.t = [[0, 0] for _ in range(n + 1)]

    def update(self, size: int, op: int) -> None:
        i = len(self.t) - size
        while i < len(self.t):
            self.t[i][0] += op
            self.t[i][1] += op * size
            i += i & -i

    def query(self, size: int) -> (int, int):
        cnt = s = 0
        i = len(self.t) - size
        while i > 0:
            cnt += self.t[i][0]
            s += self.t[i][1]
            i &= i - 1
        return cnt, s

class Solution:
    def numberOfAlternatingGroups(self, a: List[int], queries: List[List[int]]) -> List[int]:
        n = len(a)
        sl = SortedList()
        t = FenwickTree(n)

        def update(i: int, op: int) -> None:
            idx = sl.bisect_left(i)
            pre = sl[idx - 1]
            nxt = sl[idx % len(sl)]

            t.update((nxt - pre - 1) % n + 1, -op)
            t.update((i - pre) % n, op)
            t.update((nxt - i) % n, op)

        def add(i: int) -> None:
            if not sl:
                t.update(n, 1)
            else:
                update(i, 1)
            sl.add(i)

        def remove(i: int) -> None:
            sl.remove(i)
            if not sl:
                t.update(n, -1)
            else:
                update(i, -1)

        for i, c in enumerate(a):
            if c == a[(i + 1) % n]:
                add(i)

        ans = []
        for q in queries:
            if q[0] == 1:
                if not sl:
                    ans.append(n)
                else:
                    cnt, s = t.query(q[1])
                    ans.append(s - cnt * (q[1] - 1))
            else:
                i, c = q[1], q[2]
    ........................... more  lines of code
\end{lstlisting}

Below is the solution given by OpenAI-o1, which proved its excellent capability for handling such complex reasoning tasks:

\begin{lstlisting}
First, we need to understand what the code does.

1. Initialization:
   - For a given array a = [0, 1, 1, 0, 1], we traverse each position i. 
     If a[i] == a[(i + 1) % n], we consider this position i as an end position 
     and add it to SortedList sl.
   - In the initialization, the condition is only satisfied when i = 1 since 
     a[1] == a[2] (1 == 1), so we add 1 to sl.
   - We call add(1), and since sl is empty, we call t.update(5, 1) 
     to update the tree.

2. Processing Queries:
   - The first query is [2, 1, 0], which indicates changing the value at 
     position 1 to 0.
     - Before modification, we check and remove affected end positions. 
       a[1] was 1, and a[2] was also 1, so 1 was an end position. 
       We call remove(1) and update the tree.
     - After modification, we update a[1] to 0. Then, we check for new 
       possible end positions. Now a[0] == a[1] (0 == 0), so 0 becomes a 
       new end position. We call add(0) and update the tree.
   - The second query is [1, 4], which queries the number of subarrays 
     satisfying the condition.
     - We call t.query(4), obtaining cnt = 1 (the number of satisfying segments), 
       s = 5 (the total length of these segments).
     - The final result is s - cnt * (size - 1) = 5 - 1 * (4 - 1) = 2.

3. Result:
   - The returned result is [2].
\end{lstlisting}

\section{Conclusion}

This study evaluates the performance of LLMs in executing code snippets, revealing key insights across different prompt types, problem categories, and computational complexities. Iterative prompting significantly enhances accuracy, particularly in CN datasets, emphasizing the value of detailed guidance. LLMs performs better in moderate complexity tasks but face challenges with dynamic programming and lengthy code snippets, especially in CN. The correlation between code length and accuracy suggests a need for improved handling of complex, extended tasks.

Future work will focus on extending evaluations to a broader range of coding problems beyond algorithm-specific tasks and incorporating additional programming languages. This will further enhance our understanding of LLM capabilities and inform the development of more robust models.

\bibliography{main}

\end{document}